\documentclass[dvipsnames,format=sigconf]{acmart}
\usepackage[ruled,vlined,linesnumbered]{algorithm2e}
\usepackage{multirow}
\usepackage{tabularx}
\usepackage{hyperref}
\DeclareMathOperator*{\argmax}{arg\,max}
\AtBeginDocument{%
  \providecommand\BibTeX{{%
    \normalfont B\kern-0.5em{\scshape i\kern-0.25em b}\kern-0.8em\TeX}}}

\copyrightyear{2023}
\acmYear{2023}
\setcopyright{rightsretained}
\acmConference[GECCO '23 Companion]{Genetic and Evolutionary Computation Conference Companion}{July 15--19, 2023}{Lisbon, Portugal}
\acmBooktitle{Genetic and Evolutionary Computation Conference Companion (GECCO '23 Companion), July 15--19, 2023, Lisbon, Portugal}
\acmDOI{10.1145/3583133.3590581}
\acmISBN{979-8-4007-0120-7/23/07}

\newcommand{\bi}[1]{{\textcolor{blue}{#1}}}

\begin{document}

\title{Efficient Quality-Diversity Optimization through Diverse Quality Species}

\author{Ryan Wickman, Bibek Poudel, Michael Villarreal, Xiaofei Zhang, Weizi Li}
\affiliation{%
  \institution{University of Memphis}
  \city{Memphis}
  \state{Tennessee}
  \country{USA}
}
\email{{rwickman, bpoudel, tmvllrrl, xiaofei.zhang, wli}@memphis.com}

\begin{abstract}
A prevalent limitation of optimizing over a single objective is that it can be misguided, becoming trapped in local optimum. This can be rectified by Quality-Diversity (QD) algorithms, where a population of high-quality and diverse solutions to a problem is preferred. Most conventional QD approaches, for example, MAP-Elites, explicitly manage a behavioral archive where solutions are broken down into predefined niches. In this work, we show that a diverse population of solutions can be found without the limitation of needing an archive or defining the range of behaviors in advance. Instead, we break down solutions into independently evolving species and use unsupervised skill discovery to learn diverse, high-performing solutions. We show that this can be done through gradient-based mutations that take on an information theoretic perspective of jointly maximizing mutual information and performance. We propose Diverse Quality Species (DQS) as an alternative to archive-based QD algorithms. We evaluate it over several simulated robotic environments and show that it can learn a diverse set of solutions from varying species. Furthermore, our results show that DQS is more sample-efficient and performant when compared to other QD algorithms. Relevant code and hyper-parameters are available at: \bi{\underline{\url{https://github.com/rwickman/NEAT_RL}}}
\end{abstract}

\maketitle
\begin{figure}[ht]
		\centering
		\includegraphics[width=1.0\linewidth]{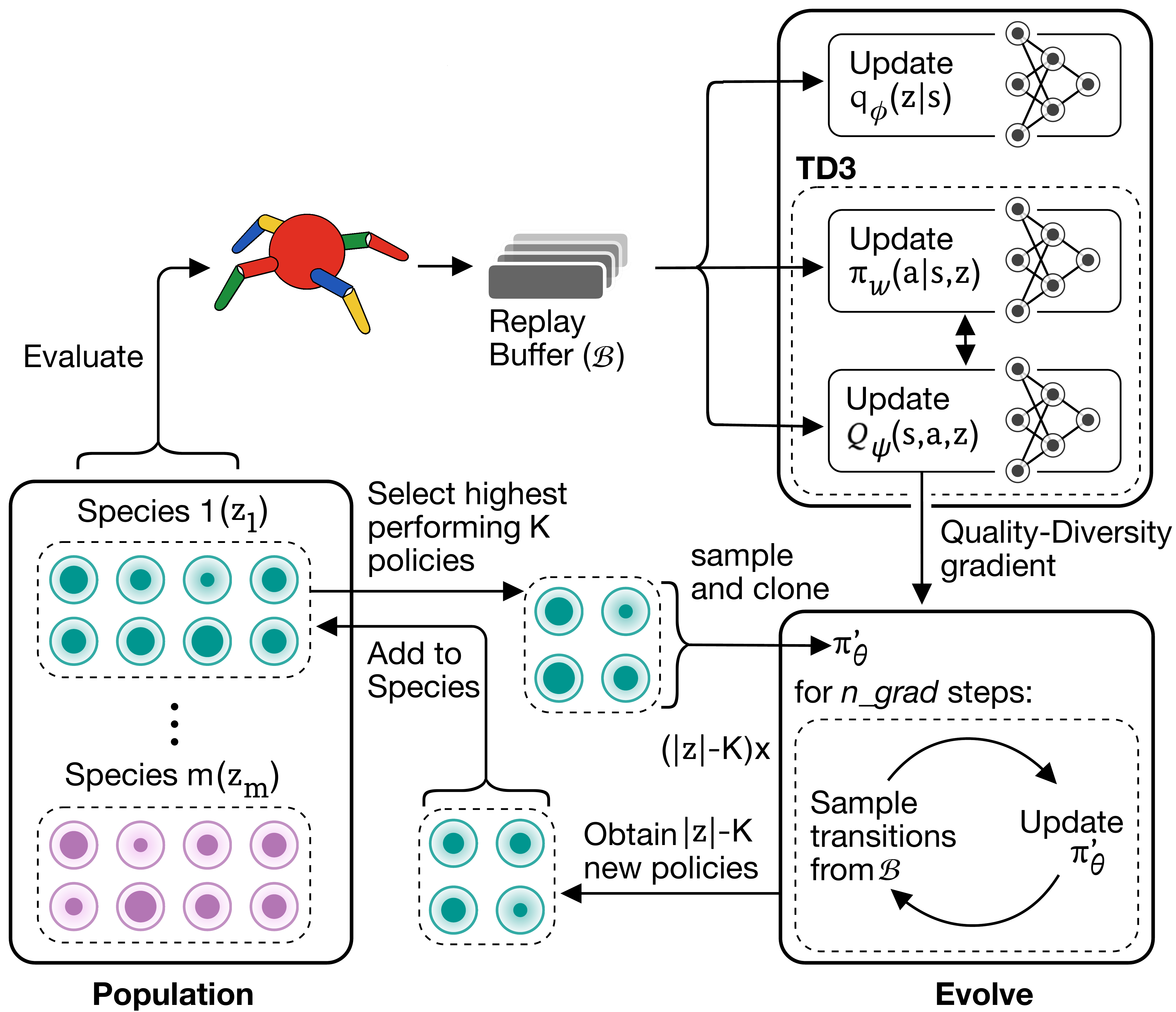}
		\caption{Diverse Quality Species (DQS) solution framework. Transitions from policy evaluations in QD environments are stored in the replay buffer $(\mathcal{B})$ and used to update the Discriminator $(q)$, Actor $(\pi)$, and Critic $(Q)$. Each species $(z)$ in the population independently evolves by selecting the $K$ highest-performing policies, updating $|z| -K$ policies with gradient-based mutations, and adding them back to the species. The Quality-Diversity gradient to jointly maximize diversity and performance is obtained from the Critic.}
		\label{fig:systematic}
\end{figure}
\section{Introduction}

Reinforcement Learning (RL) traditionally focuses on finding a single optimal policy that performs maximally on a predefined reward function. However, even without a predetermined objective, in nature we often see a variety of creative and complex behaviors emerge in different organisms that are subject to similar environments~\cite{stanley2015greatness}. This raises a question of whether a more open-minded approach, deviating from solely pursuing the reward definition, can lead to the discovery of effective solutions with greater outcomes. One such deviation is Quality-Diversity (QD) optimization~\cite{cully2017quality, pugh2016quality, chatzilygeroudis2021quality} from Evolutionary Computation (EC), where a population of diverse and high-quality solutions is preferred. Optimizing over an array of diverse solutions can prevent collapsing to local optimum and increases the likelihood of finding globally optimal solutions~\cite{lehman2011abandoning, ecoffet2021first}.

Specifically, the Novelty Search with Local Competition (NSLC) algorithm ~\cite{lehman2011abandoning} and Multi-dimensional Archive of Phenotypic Elites (MAP-Elites) algorithm~\cite{mouret2015illuminating} are two widely adopted techniques to find diverse, yet high-performing solutions. The NSLC algorithm prioritizes novelty over performance; to solve a task, it relies on encouraging solutions (organisms) within a feature neighborhood to act as differently as possible. On the other hand, MAP-Elites divides the large solution search space into a grid, with each cell corresponding to a behavioral niche and identifies the highest-performing solution (elite) in each cell. This approach provides insight on how interesting attributes of solutions are distributed within the search space and how they collectively affect performance. These methods have been applied to a variety of tasks in robotics~\cite{cully2015robots, kume2017map, lim2022dynamics}, video games~\cite{8848053, zhang2022deep}, and image generation~\cite{costa2020exploring}.

An extension of the MAP-Elites algorithm, Policy Gradient Assisted MAP-Elites (PGA-MAP-Elites)~\cite{nilsson2021policy} incorporates the ``best of both RL and EC worlds'', i.e., policy gradient operators ensures performance whereas MAP-Elites ensures exploration and diversity in solutions. By updating solutions in the direction of the noisy fitness gradient, it avoids solutions converging to a narrow peak, thus increases the likelihood of obtaining stable, high-performing solutions. Unlike MAP-Elites, PGA-MAP-Elites is not limited to low-dimensional, deterministic environments and outperforms searches that are ignorant about their objective~\cite{10.1145/3577203, pugh2015confronting}. 
Although techniques from EC demonstrate prioritizing diversity in solutions can lead to successful task completion without explicit task rewards, manual specification of metrics to measure diversity is still required. Examples include genetic distance or handcrafted and often domain-specific behavioral representations such as an archive~\cite{stanley2019designing, parker2020effective}. 

Recently, there has been an independent, but related line of work in unsupervised reinforcement learning where a growing body of work focuses on the autonomous acquisition of a diverse set of skills (latent behaviors)~\cite{Sharma2020Dynamics-Aware, kumar2020one}, or a diverse set of goals~\cite{DBLP:journals/corr/abs-1903-03698, nair2018visual, nasiriany2019planning}. These approaches leverage intrinsic motivation mechanisms~\cite{chentanez2004intrinsically} such as information gain or entropy, which drive a policy's curiosity and lead it to discover skills it finds surprising. In some works, a discriminator, conditioned on the states with an objective of maximizing mutual information between states and skills, is used as a source of intrinsic reward and encourages diversity between policies~\cite{gregor2016variational, eysenbach2018diversity}. 

In this work, we introduce a novel approach to Quality Diversity (QD) optimization called Diverse Quality Species (DQS), which combines the latest developments in QD optimization with unsupervised skill discovery. DQS re-imagines the QD optimization process by eliminating the need to explicitly define quantifiable behavioral representations traditionally found in QD algorithms. At its core, DQS partitions the solution population into different species that are independently evolved. This has several benefits, the most notable of which is a comprehensive exploration of the search space. By dividing the population into separate species, we are also able to prevent premature convergence to a single, suboptimal solution. To drive our gradient-assisted search for diverse yet high-performing solutions, we use the actor-critic algorithm Twin Delayed Deep Deterministic policy gradient (TD3)~\cite{fujimoto2018addressing}, with a key difference. Our approach conditions both the critics and the actor on the species. On one hand, this reduces the variance in value estimates of the critics without which varying degrees of performance by each species would likely make the critic's estimates noisy. While on the other hand, this allows for improved next action prediction from the actor, without relying on exact policies (constantly evolving). Further, we use a discriminator with the objective to maximize the mutual information between states and species. This encourages the various species to visit different parts of the state space i.e., encourage diversity. We evaluate our approach using four simulated robotic environments in the QD benchmarks~\cite{nilsson2021policy} (QDAnt, QDHopper, QDHalfCheetah, and, QDWalker) and compare its performance against existing state-of-the-art QD algorithms. 

In summary, our contribution is threefold:
\begin{enumerate}
\item{DQS promotes diversity between species without the need for an archive to quantify diversity or any explicit use of the behavior space. We show that the diversity generated by DQS is comparable to that of other QD algorithms.}
\item{We demonstrate that DQS can generate solutions of higher quality with state-of-the-art sample efficiency in all four environments.}  
\item{We show that DQS can generate diverse behaviors with state-of-the-art sample efficiency in QDWalker, QDAnt and QDHopper.}
\end{enumerate}

\section{Background and Related Work}
\subsection{Reinforcement Learning}
Reinforcement learning (RL) is a $T$-step episodic task where an agent with a policy $\pi_{\omega}$, parameterized by $\omega$, interacts with an environment and seeks to maximize the sum of discounted rewards, or the return $G_t = \sum_{i=t}^T \gamma^{i-t}r_i$. At each time step, the agent is given a state $s$, responds with an action $a$, and then the environment returns the next state $s'$ and reward $r$. This is formalized as a Markov Decision Process (MDP), which is represented as the tuple $(\mathcal{S}, \mathcal{A}, \mathcal{P}, \mathcal{R}, \gamma)$, where $\mathcal{S}$ represents the set of  states, $\mathcal{A}$ is the set of actions, $\mathcal{P}(s', r | s, a)$ is the environment dynamics, $\mathcal{R}(s,a)$ is the reward function, and $\gamma$ is the discount factor on the future rewards.

One state-of-the-art approach for applying reinforcement learning in a continuous action space is TD3~\cite{fujimoto2018addressing}. This algorithm trains an actor (policy) using deterministic policy gradient updates~\cite{silver2014deterministic} by directly differentiating through a critic network. The critic estimates the expected return under the policy, that is $Q_{\psi}(s,a) = \mathbb{E}_{a \sim \pi, s \sim p_{\pi}}[G_t | s, a]$. In TD3, they use a pair of critics $Q_{\psi_1}$ and $Q_{\psi_2}$ with parameters $\psi_1$ and $\psi_2$ to handle overestimation bias prevalent in the value estimates. To stabilize training updates, they use a pair of target critics and a target actor with parameters $\psi_1'$, $\psi_2'$, and $\omega'$. These target networks are updated by slowing interpolating with a value of $\tau$  between the targets and the current parameters. Furthermore, the target and policy updates are delayed by only performing them every $d$ critic update steps.

\subsection{Quality-Diversity (QD)}
In QD, the goal is to produce a population of diverse and high-quality solutions on a predefined task. Diversity is measured through the range of behaviors produced by the solutions. Behaviors are typically defined in an environment as the Behavior Descriptor (BD)~\cite{pugh2016quality} with the full set of BDs is referred to as the \emph{behavior space}~\cite{mouret2015illuminating}. The BD is task-dependent. For example, it could be the final position of the agent~\cite{lehman2011abandoning, pierrot2022diversity} or the proportion of time the robot's feet contact with the ground~\cite{cully2015robots, nilsson2021policy}.    

The objective of a QD algorithm is to find a population of policies (or solutions) $\Pi$ that maximizes:
\begin{equation}
    \Pi^* = \argmax_\Pi \mbox{QD-Score}(\Pi)
\end{equation}
where QD-Score is the sum of fitness scores over all policies in the population $\pi \in \Pi$. That is $\sum_{\pi \in \Pi}f(\pi)$, where $f(\cdot)$ is the fitness function that returns the fitness for a given policy. The fitness measures the quality of a solution. In our experiments, it will be defined as the sum of non-discounted rewards from the environment.

\subsection{Related Work}

One of the main works that motivates our approach is PGA-MAP-Elites~\cite{nilsson2021policy}, which introduces a policy gradient (PG) variation operator that is used in our work. Similar to other QD-related works~\cite{kume2017map, lim2022dynamics, fontaine2020covariance, colas2020scaling, pierrot2022diversity}, they utilize the MAP-Elites algorithm~\cite{mouret2015illuminating} to store the solutions in an archive. One disadvantage of MAP-Elites is it requires a statically generated archive that must be designed before solving the task. We circumvent this necessity by encouraging diversity through speciation and an auxiliary diversity reward. Additionally, we avoid the directional genetic variation operator~\cite{vassiliades2018discovering} they employ.

There have been numerous previous studies that approached RL from an information-theoretic perspective~\cite{ziebart2008maximum, eysenbach2018diversity, schulman2017equivalence, haarnoja2018soft, DBLP:journals/corr/abs-1903-03698}. For example, maximizing the mutual information between states and goals/skills~\cite{ DBLP:journals/corr/abs-1903-03698, eysenbach2018diversity} or by utilizing a dynamics model to measure the predictability of the policy~\cite{houthooft2016vime, Sharma2020Dynamics-Aware}. Among these prior studies, we drew inspiration from Diversity is All You Need (DIAYN)~\cite{eysenbach2018diversity}, in which a single policy is trained to acquire multiple skills. A \emph{skill} serves as a latent identifier for the policy, leading to a substantial change in its behavior. In their work, they search for a set of skills that are as diverse as possible, such that the skills are distinguishable by the states that they visit. They do not train the skill-policy on environment reward; instead, they derive a diversity reward from a discriminator that predicts the skill given a state. While we use a variant of the diversity reward in this work, we don't train a single skill-conditioned policy that only focuses on diversity. As we are focused on maximizing the QD objective, we train a speciated population of policies that aim to maximize quality and diversity. Thus, we do use the environment reward in our objective. While other work combines diversity and environment rewards~\cite{kumar2020one}, their focus is on a single skill-conditioned policy that only uses the diversity reward once a minimum return has been achieved by the skill. On the contrary, we apply the diversity reward across the entire training process.

Prior studies have demonstrated that speciation is an effective means of maintaining diversity and fostering independently varied solutions\cite{goldberg1987genetic, stanley2002evolving, lehman2011abandoning, martins2020applying}. In this particular scenario, the competition is localized across distinct niches, thereby maintaining a diverse array of elite solutions. Speciation has been utilized to propel evolution towards more optimal solutions by implementing explicit fitness sharing~\cite{goldberg1987genetic}, where high-performing species will have a higher reproductive rate compared to those with lower performance.   In the Neuroevolution of Augmenting Topologies (NEAT) algorithm~\cite{stanley2002evolving}, speciation is performed in every generation based on the topological distance calculated by a compatibility function. In our work, we maintain a constant number of species and maintain the solutions (or their offspring) within each species throughout the entire training process. 

The work most similar to ours is Quality-Diversity Policy-Gradient (QD-PG)~\cite{pierrot2022diversity} where they perform policy gradient updates similar to PGA-MAP-Elites. However, they define an additional diversity policy gradient based on the distances between neighboring solutions in the behavior space. Unlike QD-PG, we do not utilize MAP-Elites, define an additional diversity critic to perform diversity updates, or utilize a predefined behavior space to encourage diversity. Our approach conditions the critic on the species and trains it on a joint reward function that incentivizes both high-performing and diverse solutions relative to other species.


\section{Diverse Quality Species}
In Figure~\ref{fig:systematic}, we present a high-level overview of our approach, which integrates the concept of QD with unsupervised skill discovery, referred to as Diverse Quality Species (DQS). Further details of the methodology can be found in Algorithm~\ref{alg:dqs}. The approach involves evaluating a set of species in an environment, with transitions being stored in a replay buffer. These transitions are utilized to update the species critic, species actor, and discriminator. The role of the discriminator is to generate a species diversity reward, which promotes distinct behaviors among the different species. The species critic generates a Quality-Diversity gradient, which is employed to drive the evolution of the population and produce a new generation of policies from the top K elites within each species.

In the rest of this section, we first describe how we maximize diversity amongst species (Section~\ref{encouraging-diversity}). Then, we discuss how we trade-off quality and diversity in our approach (Section~\ref{qd-tradeoff}), followed by the discussion on how we train our critics to produce the policy gradients for our population (Section~\ref{training-critic}). Lastly, we discuss how we evolve the population between successive generations (Section~\ref{sec:evolve_pop}).

\subsection{Encouraging Diverse Species}
\label{encouraging-diversity}
Speciation is one of the key components of our work. It allows us to independently drive the evolution of each species and shape their distinct behaviors. This enables us to cultivate targeted innovations and make precise improvements at a granular level. This intuition leads to our ultimate goal: we want to train a robust set of high-performing species $~\mathbf{Z}$ where each species $z$ has maximum diversity. Each species is a collection of deep neural network policies $\pi_{\theta_j} \in z$, where each policy is parameterized by a unique set of parameters $\theta_j$. Thus, the union over all policies in each species will make up our population $\Pi = \bigcup_{\mathbf{Z}} z$.

With the aim of encouraging diversity within our population, we adopt a framework that emphasizes promoting diversity among species. Specifically, we believe that enforcing diverse state visitation distributions among species will enable comprehensive coverage of the behavior space. Similar to DIAYN~\cite{eysenbach2018diversity}, where instead of skills we have species, this can be framed as maximizing the mutual information between the set of states and species, $I(\mathbf{S}; \mathbf{Z})$. 

In short, the diversity objective we want to maximize is:
\begin{equation}
\begin{split}
    \mathcal{F}(\theta) &= I(\mathbf{S}; \mathbf{Z}) \\
    &=  -\mathcal{H}(\mathbf{Z}|\mathbf{S}) + \mathcal{H}(\mathbf{Z})
\end{split}
\end{equation}
The entropy over the set of species $\mathcal{H}(\mathbf{Z})$ will be maximized if the number of policies in each species are equal. As such, we enforce all species to contain an equal number of policies. The conditional entropy of the species given the states $\mathcal{H}(\mathbf{Z}|\mathbf{S})$ is minimized when there is full coverage over the state space and each species visits a different subset of states. Alternatively, this can be viewed as maximizing the conditional probability of the species given the states $p(z|s)$. Maximizing this conditional likelihood indicates a desire for the species to be distinguishable by the states they visit. 

As it is intractable to integrate over all the states and species to compute $p(z|s)$ exactly, we approximate it using a neural network $q_{\phi}(z|s)$. Following prior work~\cite{10.5555/2981345.2981371}, we introduce a lower bound over the mutual information objective:
\begin{equation} \label{eq:2}
\begin{split}
    \mathcal{F}(\theta) &=  \mathbb{E}_{z \sim p(z), s \sim \pi(z)}\left[\log\frac{p(z|s)}{ p(z)}\right] \\
    &=  \mathbb{E}_{z \sim p(z), s \sim \pi(z)}\left[\log\frac{q_{\phi}(z|s)}{p(z)}\right] 
    + \mathcal{D}_{KL}(p(z|s) \parallel q_{\phi}(z|s))\\
    &\geq  \mathbb{E}_{z \sim p(z), s \sim \pi(z)}\left[\log\frac{q_{\phi}(z|s)}{p(z)}\right]
\end{split}
\end{equation}
where $\mathcal{D}_{KL}$ denotes the Kullback-Leibler divergence. This objective is tightest when $q_{\phi}(z|s) = p(z|s)$. Thus, this bound can be improved by performing maximum likelihood estimation over the parameters $\phi$ of the approximate function $q_{\phi}(z|s)$. We refer to this function as the discriminator.

The discriminator $q_{\phi}(z|s)$ will differentiate between the various species based on the states visited by their respective policies. Intuitively, we use this as a way to encourage different species to behave differently in the environment, such that the states they visit can be differentiated by this discriminator.

\subsection{Quality-Diversity Trade-off}
\label{qd-tradeoff}
In many environments, if only diversity is pursued, only a small subset of the diverse behaviors will be useful for the downstream task~\cite{eysenbach2018diversity, Sharma2020Dynamics-Aware}. As we want to increase the overall population quality, we need to maximize the cumulative reward received by the environment from each policy. This forms a trade-off between quality and diversity. We want the individual species to perform well at the target task, but behave differently. 


 As we want to maximize the diversity objective in Equation~\ref{eq:2}, we define a species diversity reward:
 \begin{equation}
r_{z} = \log q_{\phi}(z|s) - \log p(z)
 \end{equation}
Then, we formulate a QD reward as the combination of the environment reward plus the discounted species diversity reward:
\begin{equation}
    r_{qd} = r + \lambda r_{z}
\end{equation}
where $r$ denotes the environment reward and $\lambda$ weights the importance of maximizing the diversity of species. This composite reward function serves as the primary objective for our policies. The magnitude of the diversity reward is controlled through the value of $\lambda$, enabling the trade-off between diversity and quality to be easily adjusted. For example, a value of $\lambda = 0$ prioritizes the quality of the population over the diversity of species. 

\subsection{Training Species Critic and Actor}
\label{training-critic}
We use the TD3 algorithm to train a pair of critic neural networks $Q_{\psi_1}$ and $Q_{\psi_2}$ to approximate the action-value function. In our case, they are trained to maximize $r_{qd}$. Additionally, as different species can have varying levels of task performance, we can reduce the estimation variance by conditioning the critics on the species (e.g., $Q_{\psi_1}(s, a, z)$). We refer to these species-conditioned critics as \emph{species critics}. By conditioning the critic on the species, we allow for species-level predictions to be made. This approach is particularly useful in cases where multiple species are evaluated over the same state. In these scenarios, it is expected that each species will exhibit a different true expected return, due to differences in the diversity reward for a given state-species pair. In the extreme case, this diversity reward is highest for the most likely species, as indicated by a probability of 1.0 from the discriminator, and is minimized for all other species.

\begin{algorithm}[t]
    \SetAlgoNoLine
    \SetKwProg{Fn}{Function}{}{end}
    \SetKwFunction{FMain}{DQS}
    \SetKwFunction{FEvaluate}{Evaluate($\Pi$)}
    \SetKwFunction{FEvolve}{Evolve($\Pi$)}
    \SetKw{In}{in}
    
    Initialize $Q_{\psi_1}, Q_{\psi_2}, \pi_{\omega}$, and $q_{\phi}$\;
    Initialize targets $Q_{\psi_1'}, Q_{\psi_2'}$, and $\pi_{\omega'}$\;
    Initialize population $\Pi$\;
    Initialize species $\mathbf{Z}$ by equally dividing $\Pi$ into $m$ species\;
    Initialize replay buffer $\mathcal{B}$\;

    \Fn{\FMain}{
        \While{$i < num\_eval$}{
            Evaluate($\Pi$) \tcp*{Evaluate entire population}
            $\Pi =$ Evolve($\Pi$) \tcp*{Generate new population}
            $i \mathrel{+}= |\Pi|$\;
        }
        \Return $\Pi$ \; 
    }

    \Fn{\FEvaluate}
    {
        \ForEach{$\pi_{\theta}$ \In $\Pi$}
        {
            \For{$t = 1$ \KwTo $T$}
            {
                Sample action $a \sim \pi_{\theta}(s)$\;
                Take action and observe $s', r \sim \mathcal{P}(r, s' | s, a)$\;
                $r_z = \log q_{\phi}(z|s') - \log p(z)$\;
                Store $(s, a, r, r_z, s', z)$ in $\mathcal{B}$\;
                \If{$t \mod critic\_update\_freq$}{ \label{alg:update_critic}
                    Update $Q_{\psi_1}, Q_{\psi_2}, \pi_{\omega}$ with TD3\;
                    Update $\phi$ to maximize $q_{\phi}(z|s)$ using SGD\; 
                }\label{alg:update_critic_end}
            }
            Update average fitness of $\pi_{\theta}$\;
        }   
    }

    \Fn{\FEvolve}
    {
        $\Pi' = \{\}$\; 
        \ForEach{$z$ \In $\mathbf{Z}$}{
            $S =$ Select highest performing $K$ policies from $z$\;
            \For{$k = 1$ \KwTo $|z| - K$}{
                $\pi_{\theta}' =$ randomly sample and clone from $S$\;
                \For{$l = 1$ \KwTo n\_grad}{
                    Sample $N$ states by $z$ from $\mathcal{B}$\;
                    $\nabla_{\theta} J(\theta$) = $\frac{1}{N} \sum \nabla_\theta \pi'_\theta(s) \nabla_a Q_{\psi}(s, a, z) |_{a=\pi'_{\theta}(s)}$\;
                }
                Add $\pi'_{\theta}$ to $z$\;
            }
            Add $z$ to $\Pi'$
        }
        \Return $\Pi'$        
    }

\caption{Diverse Quality Species (DQS) Algorithm.}
\label{alg:dqs}
\end{algorithm}

In TD3, the next action needs to be sampled by the policy to perform critic updates. However, we are constantly evolving the population and thus a policy that generated trajectories in the replay buffer $\mathcal{B}$ may terminate before they are sampled for training. To rectify this issue, we train a \emph{species actor} $\pi_{\omega}(a | s, z)$ that is conditioned on the current state and a species. This species actor is not evaluated in the environment but is trained on trajectories sampled from the replay buffer through deterministic policy gradient updates. This allows us to perform critic updates without concerning ourselves with what exact policy generated the trajectories.

Every $critic\_update\_freq$ steps in the environment, we perform a singular update on the species critic, species actor, and discriminator. These updates can be seen in Algorithm~\ref{alg:dqs} on Lines~\ref{alg:update_critic}-\ref{alg:update_critic_end}. The species critic's target is produced using the species actor as follows:
\begin{equation}
    y = r_{qd} + \gamma \min\limits_{i=1,2} Q_{\psi'_i}(s', \pi_{\omega}(a' | s', z) + \epsilon)
\end{equation}
where $\epsilon \sim \mbox{clip}(\mathcal{N}(0, \sigma), -c, c)$ is normally distributed noise with variance $\sigma$ clipped by $c$~\cite{fujimoto2018addressing}. Adding noise to the action estimate is a standard procedure in TD3. It allows for the smoothing of action-value peaks and improved stability in the learning process. In our implementation, the species actor serves as a proxy for the overall population of diverse species, as demonstrated in the species critic's target.

\subsection{Evolving the Population}
\label{sec:evolve_pop}
As previously stated, we want to maximize $\mathcal{H}(\mathbf{Z})$, so we initialize the population by equally distributing the policies, with randomly initialized parameters, into $m$ species. Unlike the species actor, these policies are not conditioned on their species, but only on the current state $\pi_{\theta}(a|s)$. We will keep these species and their length $|z|$ fixed throughout the training process, and only replace solutions with offspring from the same species.

In many recent QD methods, we have noted a common trend where evaluation is performed only on new policies that have been generated from previous evolution steps~\cite{nilsson2021policy, pierrot2022diversity}. However, in stochastic environments, a policy may produce a high fitness value by chance during its initial evaluation, leading to several iterations of non-evolution and difficulty in replacing these ``lucky'' policies. To address this issue, we evaluate the entire population at every iteration, maintaining an average of each policy's fitness values obtained from the environment. This, we believe, provides a more accurate representation of the overall quality of the policies and leads to a more effective evolution process. We describe the full evaluation procedure in Algorithm~\ref{alg:dqs} in the function Evaluate.

After the entire population is evaluated, we evolve the population $\Pi$ to produce a new population $\Pi'$, as shown in Algorithm~\ref{alg:dqs} in the function Evolve. We begin by selecting the highest performing K policies (the elites) from each species and discarding the rest of the policies. This is done to ensure only the most promising candidates are used to produce the next generation. For each species, we sample $|z| - K$ random policies from the current species, run $n\_grad$ QD policy gradient updates on them, and add them as new policies to the species. When sampling trajectories for policy gradient updates, we only sample trajectories that are produced by their corresponding species. This way we further encourage diversity in our population by training them on different subsets of the replay buffer.

\begin{table}
\begin{center}
    \begin{tabular}{lc}
        \toprule
        Hyperparameter & Value \\
        \hline
        Population Size & 64 \\
        Number of Species ($m$) &  8 \\
        Diversity Reward Scale ($\lambda$) & 0.05 \\
        Species Elites Value ($K$) & 4 \\
        Policy Update Steps ($n\_grad$) & 64 \\
        Critic Update Freq. ($critic\_update\_freq$) & 8 \\
        Policy Hidden Size & 128 \\
        Species Actor Hidden Size & 256 \\
        Species Critic Hidden Size & 256 \\
        Discriminator Hidden Size & 256 \\
        Species Actor/Critic and Discriminator Learning Rate & 0.003 \\
        Policy Learning Rate & 0.006 \\
        Number of Evaluations ($num\_eval$) & $10^{5}$ \\
        Batch Size ($N$) & 256 \\ 
        Discount Factor ($\gamma$) & 0.99 \\
        Species Target Update Rate ($\tau$) & 0.005 \\
        TD3 Exploration Noise  & 0.2 \\
        TD3 Smoothing Variance ($\sigma$) & 0.2 \\
        TD3 Noise Clip ($c$) & 0.5 \\
        TD3 Target Update Freq. ($d$) & 2 \\
        Replay Buffer Size & $2^{19}$ \\
        \bottomrule
 \end{tabular}
\end{center}

\caption{Hyperparameters for the DQS algorithm.}
\label{tab:hyper}
\end{table}

\begin{figure*}[t]
		\centering
		\includegraphics[width=1.0\linewidth]{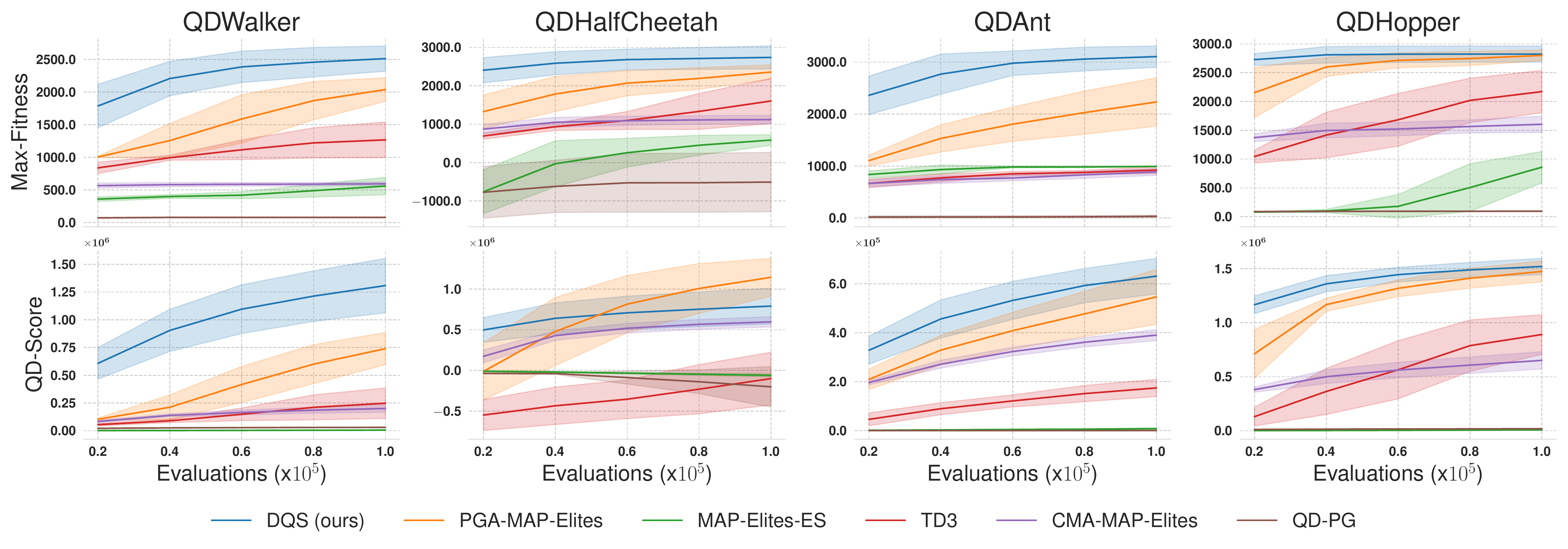}
		\caption{Results over the QD-Gym environments where each axis denotes the proportion of time each foot is on the ground. Each task and algorithm was repeated over 10 random seeds for a total of 100k evaluations. For each evaluation metric, we plot the average and show the standard deviation through the shaded region. At 100K evaluations, DQS outperforms all algorithms across all four environments in Max-Fitness and outperforms all algorithms across all environments except QDHalfCheetah in QD-Score.}
		\label{fig:exp_results}
\end{figure*}

\section{Experiments}
In this section, we show that DQS is more sample-efficient than other QD approaches by showing it can reach better quality performance quicker and comparable diversity when measured over the total number of environment evaluations.
\subsection{Experimental Setup}
We test on 4 environments from QD-Gym~\cite{nilsson2021policy}: QDWalker, QDHalfCheetah, QDAnt, and QDHopper.
Each of these tasks involves a simulated robot that needs to learn an energy-efficient way to walk. In these environments the states are defined as the current center of gravity height, x, y, and z velocity, roll, pitch and yaw angles, and the relative positions of the robot's joints. Actions are continuous-valued torques on each of their respective rotors. As one goal in QD is to increase the diversity of the population, the BD is set up to measure different walking behaviors in the solutions. The BD of each task is the proportion of time each foot of the robot is in contact with the ground. The dimensions of these are 2, 2, 4, and 1, respectively. 

In all experiments, we use a population size of 64, 8 species, and set $\lambda = 0.05$. As we equally divide the population into $m$ species, the species length $|z|$ is fixed at 8 for all species. We use a total of 3 layers for all networks and use the Adam optimizer~\cite{kingma2014adam}. The full set of hyperparameters can be found in Table~\ref{tab:hyper}. As we want to measure the sample-efficiency of our approach, we evaluate all algorithms over 10 random seeds for a maximum of 100k evaluations.

\subsection{Baselines and Evaluation Metrics}
We evaluate DQS against various baselines: PGA-MAP-Elites~\cite{nilsson2021policy}, CMA-MAP-Elites~\cite{fontaine2020covariance}, MAP-Elites-ES~\cite{colas2020scaling}, QD-PG~\cite{pierrot2022diversity}, and TD3~\cite{fujimoto2018addressing} with a CVT-MAP-Elites archive~\cite{vassiliades2017using}. The results of these baselines were graciously provided to us by the authors of \emph{Empirical analysis of PGA-MAP-Elites for Neuroevolution in Uncertain Domains}~\cite{10.1145/3577203}, where we used the first 10 runs of each method for displaying the results.


In order to evaluate the approaches, we define two metrics:
\begin{itemize}
    \item \textbf{QD-Score}: the sum of all fitness in the archive. In our setting, this is equivalent to the sum of all non-discounted returns in the archive.
    \item \textbf{Max-Fitness}: the maximum return across all solutions in the archive.
\end{itemize}

In conventional Quality-Diversity (QD) methods, a fixed archive is typically utilized to calculate the QD-Score for all benchmark approaches. However, our approach differs in that we do not maintain an archive for our population. Instead, for comparison purposes, we utilize the same CVT-MAP-Elites archive as PGA-MAP-Elites to store the fitness values of our solutions generated throughout the training process. After each batch of evaluations over the population, we update the archive with copies of our solutions, but do not utilize the solutions in those cells. This can be seen as having an archived population that can be accessed if necessary, but not actively utilized in subsequent evolution cycles.


\subsection{Experimental Results}
In Figure~\ref{fig:exp_results}, we display the experimental results over the QD-Gym environments. The average performance is depicted, with the shaded region representing the standard deviation for each metric. Our proposed algorithm, DQS, demonstrates superior Max-Fitness and QD-Score in all environments, with the exception of QDHalfCheetah on QD-Score. Notably, the QDWalker and QDAnt environments exhibit the most significant improvement compared to previous methods, as evidenced by both Max-Fitness and QD-Score metrics.

\begin{figure*}[t]
		\centering
		\includegraphics[width=1.0\linewidth]{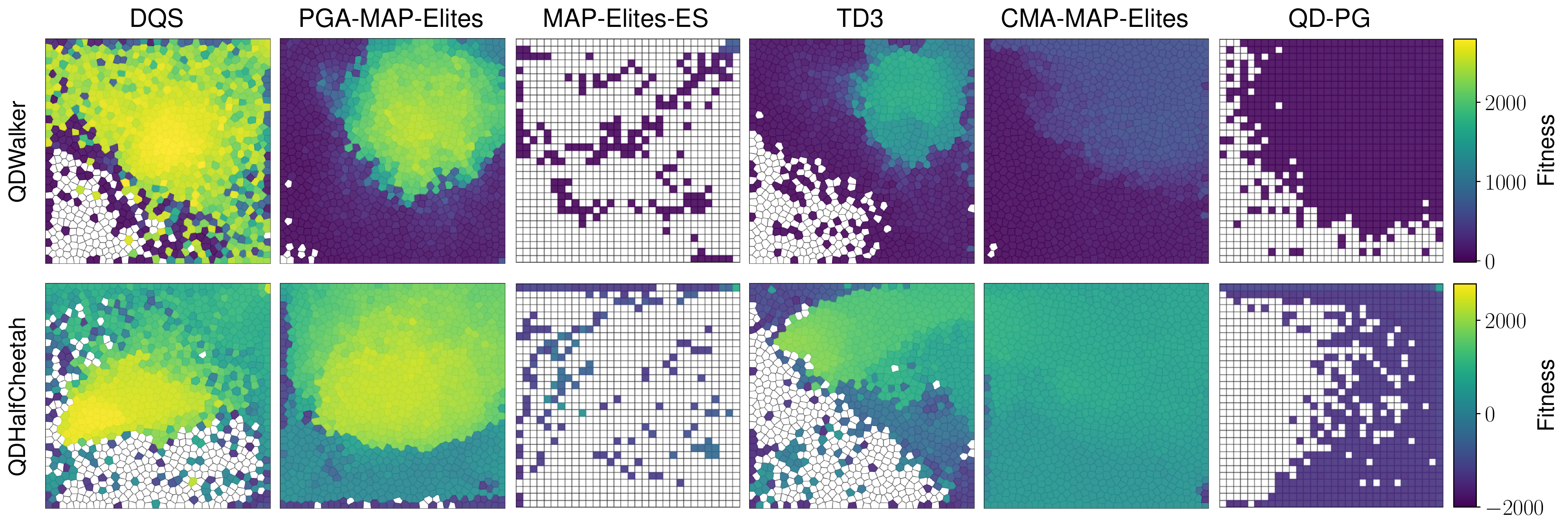}
		\caption{A sample of the archives found by each algorithm in the QDWalker and QDHalfCheetah environments. DQS has higher quality solutions, but lower coverage compared to PGA-MAP-Elites and CMA-MAP-Elites in both the environments. However, DQS demonstrates dispersion of high quality behaviors throughout the behavior space in the QDWalker environment.}
		\label{fig:archive}
\end{figure*}

In Figure~\ref{fig:archive}, we plot a sample of the archives found by each algorithm in the QDWalker and QDHalfCheetah environments, as these both have 2-dimensional BDs. Although DQS consistently produces higher quality solutions in most cells, its overall coverage falls short when compared to PGA-MAP-Elites and CMA-MAP-Elites algorithms. This discrepancy is likely due to the predominance of low-quality solutions in those unpopulated cells.  Thus, discovering such behaviors through policy-gradient updates, which focus on maximizing reward, may prove challenging. The results of the TD3 archive support this hypothesis, as its coverage is similar, albeit inferior, to our approach. Furthermore, when maximizing quality without an explicit archive, these regions of low-quality behaviors are likely to be under-explored.

The QDWalker environment reveals a noteworthy characteristic of DQS: the distribution of high-quality behaviors appears to be more dispersed. Unlike PGA-MAP-Elites, which displays a significant cluster of high-quality behaviors, DQS discovers high-quality behaviors throughout the behavior space. This may suggest that our approach to foster diversity drives policies to learn high-quality solutions for the behaviors they have been assigned, rather than directing them towards previously identified regions of high-quality behavior.

\begin{figure}[h]
		\centering
		\includegraphics[width=1.0\linewidth]{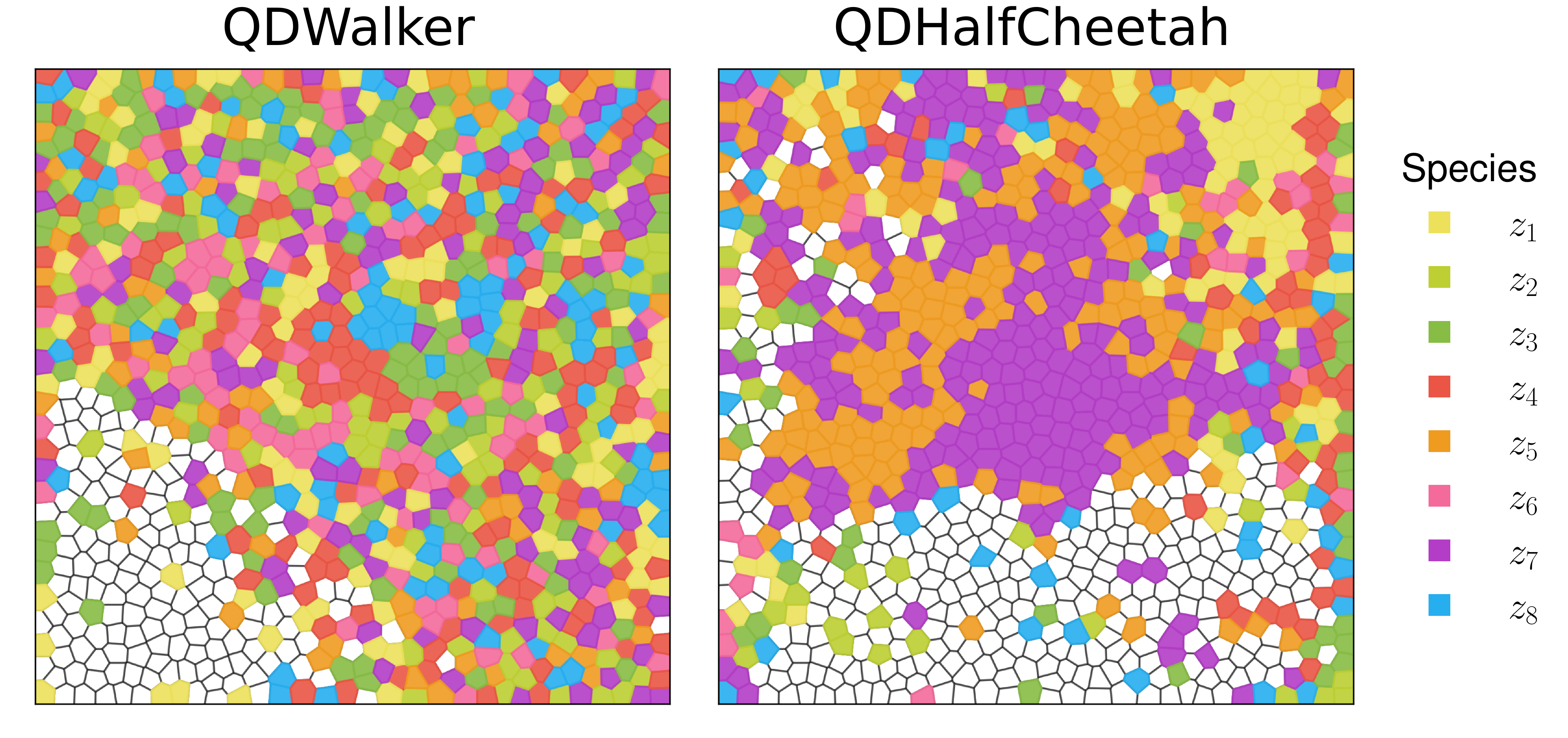}
		\caption{Archive of diverse behaviors discovered by DQS for each species in the QDWalker and QDHalfCheetah environments. Distinct species clusters can be seen in QDHalfCheetah environment, most notably for species $z_5$ and $z_7$ whereas for the QDWalker environment all species are more distributed throughout the behavior space.}
		\label{fig:species_archive}
\end{figure}

\subsection{Analyzing Species Diversity}
In Figure~\ref{fig:species_archive}, we show DQS's ability to produce species with diverse behaviors for the QDWalker (left) and QDHalfCheetah (right) environments. For QDWalker, the behavior niches learned by each of the 8 species are well-distributed within the behavior space. Three species, $z_3$, $z_4$, and $z_8$, form bordering clusters in the area of highest performance for QDWalker, but still learn behaviors all throughout the behavior space attesting to DQS's diversity ability. In the QDHalfCheetah environment, the species' behaviors form more distinct species clusters in the behavior space. Specifically, two dominating clusters form by species $z_5$ and $z_7$. For both environments, these dominating clusters form in the behavior space area corresponding to the highest fitness, while also exploring other parts of the behavior space. Another distinct species cluster with above average performance (corresponding to $z_1$) forms in the behavior space's top right. One notion about cluster formation is they lead to good performance; however, this is not necessarily the case as QDWalker illustrates species can diversify and find high-performing behaviors without clustering together. For example, species $z_7$, $z_2$, and $z_4$ find high-performing behaviors in covered corners of the behavior space without individually clustered as a species.

As QDHalfCheetah is the only environment without state-of-the-art QD-Score performance, this leads us to further investigate the cause. The two main clusters in Figure~\ref{fig:species_archive} for QDHalfCheetah represent the species that learn to walk during training, while the remaining species merely learn to stand efficiently; the species that learn efficient standing are stuck in a local optimum. This is because these species are not properly evolving, thus, remaining stagnate. Table~\ref{tab:species_age} illustrates this fact where we present the average age (i.e., the number of evolutions without being replaced) and the average fitness of the elites in each species. In the QDHalfCheetah environment, the average age of elites is negatively correlated with average fitness. Species $z_5$ and $z_7$, the dominating clusters, have the lowest average ages, which is evidence they are not stuck in a local optimum. The other species, however, have much higher average ages; for example, $z_1$ and $z_8$ have average ages of 297 and 234, respectively, while performing worse compared to $z_5$ and $z_7$. As species $z_5$ and $z_7$ largely outperform other species, fewer species are competing for those high fitness behaviors. This explains why these species behaviors are easy to identify in the low-dimensional behavior space. In the QDWalker environment, there is no such negative correlation between average species age and average species fitness.

With traditional RL techniques, stagnation is not as prevalent as an issue as policies (solutions) are frequently updated; however, with DQS, only 64 policy gradient updates are performed to a randomly sampled elite. For QDHalfCheetah, these 64 policy gradient updates are not always enough to escape the local optimum, hindering the removal of  suboptimal solutions from the population. This reveals an important direction that we will explore in future work: preventing the stagnation of species.


\begin{table}
\begin{center}
\begin{small}
    \begin{tabularx}{\columnwidth}{lXXXX}
    \toprule
    \multirow{2}{*}{Species} & \multicolumn{2}{c}{QDHalfCheetah} & \multicolumn{2}{c}{QDWalker}  \\
    \cline{2-5} & Avg. Age & Avg. Fitness & Avg. Age & Avg. Fitness \\ 
    \hline
    $z_1$ & 297 & 899 & 13.25 & 2306.86 \\
    $z_2$ & 46 & 901.11 & 9.25 & 1872.96 \\
    $z_3$ & 87 & 955 & 5.75 & 2334.15 \\
    $z_4$ & 219 & 885.41 & 2.75 & 2149.50 \\
    $z_5$ & 3.50 & 2561.22 & 28.25 & 2081.14 \\
    $z_6$ & 43 & 934 & 2.25 & 2277.59 \\
    $z_7$ & 11.25 & 2577.46 & 12.50 & 2184.77 \\
    $z_8$ & 234 & 871.73 & 25.75 & 2358.89 \\
    \bottomrule
 \end{tabularx}
\end{small}
\end{center}
\caption{Average age and fitness statistics of elites in each species for QDHalfCheetah and QDWalker.}
\label{tab:species_age}
\end{table}

\subsection{Ablation Studies}
We conduct ablation studies on the QDWalker and QDHopper environments to test the effect of speciation and species diversity. We test against two different scenarios: removing speciation (i.e., setting $m = 1$), and preserving speciation, but removing the species diversity bonus (i.e., setting $\lambda = 0$). We run each for a total of five runs, $100$k evaluation, and average the results over Max-Fitness and QD-Score. We present our results in Table~\ref{tab:ablation}.

The speciation and species diversity reward strategies are initially utilized with the objective of enhancing population diversity. However, our experimental results demonstrate that they have a significant impact on solution quality as well, as evidenced by the observed increase in both Max-Fitness and QD-Score when both strategies are employed.  Particularly in the QDWalker task, the QD-Score improves nearly two-fold and Max-Fitness experiences at least a 55\% increase. While the results in the QDHopper task are not as dramatic, there is still evidence that speciation and diversity reward have an impact on improvement. While the effect of each strategy may vary between environments, our findings suggest that the integration of speciation and species diversity can lead to both increased population diversity and improved solution quality.




\begin{table}
\begin{center}
\begin{small}
    \begin{tabularx}{\columnwidth}{lXXXX}
        \toprule
        \multirow{2}{*}{Method} & \multicolumn{2}{c}{QDHopper} & \multicolumn{2}{c}{QDWalker}  \\
        \cline{2-5} & Max-Fitness & QD-Score & Max-Fitness & QD-Score \\ 
        \hline
        DQS & $\mathbf{2823.49}$ & $\mathbf{1.52 \times10^6}$ & $\mathbf{2617.91}$ & $\mathbf{1.31 \times10^6}$ \\
        DQS ($\lambda = 0)$ & $2780.18$ & $1.45 \times10^6$ & $1639.82$ & $7.97 \times10^5$\\
        DQS ($m = 1)$ & $2750.91$ & $1.35 \times10^6$ & $1687.92$ & $8.78 \times10^5$ \\
        \bottomrule
 \end{tabularx}
\end{small}
\end{center}

\caption{Ablation experiments results, averaged over 5 random seeds for 100k evaluations. We can see that both speciation and the species diversity reward have an effect on both Max-Fitness and QD-Score, where it is highest when both are used.}
\label{tab:ablation}
\end{table}

\section{Conclusion} 
 In this work, we introduce Diverse Quality Species (DQS), a novel approach to training a diverse and high-performing population of solutions without the need for explicit behavioral representations. This is an alternative to traditional Quality Diversity (QD) algorithms, and our results demonstrate that DQS outperforms existing methods by producing higher quality solutions with comparable diversity. 
 DQS leverages speciation in the population and enforces diverse state visitation through a discriminator that approximates a diversity objective (by maximizing the mutual information between states and species). This leads to the emergence of distinct behaviors among species, resulting in a thorough exploration of the search space. Our training process balances the interplay between quality and diversity by combining both the environment reward and species diversity reward, with a scalar that can be adjusted to tune the trade-off.

We employ the actor-critic algorithm Twin Delayed Deep Deterministic policy gradient (TD3) with modifications to reduce variance in the critic's value estimation. Based on comparisons with existing QD algorithms, our experiments demonstrate that DQS achieves state-of-the-art sample efficiency in generating high-quality solutions across four robotic environments, as well as in generating diverse behaviors among species. We have also conducted ablation studies to evaluate the impact of our algorithmic design decisions, such as the choice of speciation and the scalar balancing the quality-diversity trade-off. Our results indicate that without speciation and solely relying on the environment reward, both quality and diversity suffer. Further, the formation of clusters of species in the behavior space supports our enforced diverse state visitation among species. Although our approach does not maintain an archive, we compared our solutions to five existing QD algorithms in the QDHalfCheetah and QDWalker environments to assess coverage. While DQS does not have the same coverage as PGA-MAP-Elites or CMA-MAP Elites, this is likely due to the solutions not explored by DQS corresponding to low-quality behaviors.
 
In the future, we plan to investigate the impact of recent findings such as the elimination of primacy bias~\cite{nikishin2022primacy} and the sample efficiency offered by network resets~\cite{d2022sample} on the evolution of species. We aim to further improve the coverage of the behavior space and make our approach even more sample efficient through the use of dynamics models~\cite{lim2022dynamics}. Additionally, we will examine the time efficiency offered by our algorithm, and explore the potential for parallelism offered by our independent evolution of species.

\bibliographystyle{ACM-Reference-Format}
\bibliography{reference}

\appendix

\end{document}